\documentclass[10pt,twocolumn]{article}

\usepackage[letterpaper,top=0.75in,bottom=0.9in,left=0.62in,right=0.62in,columnsep=0.22in]{geometry}
\usepackage{mathptmx}            % Times for text + math
\usepackage[T1]{fontenc}
\usepackage[utf8]{inputenc}
\usepackage{graphicx}
\usepackage{amsmath,amssymb}
\usepackage{booktabs}
\usepackage{caption}
\usepackage{titlesec}
\usepackage{enumitem}
\usepackage{stfloats}
\usepackage{balance}
\usepackage[hidelinks]{hyperref}
\usepackage{url}

\titleformat{\section}{\centering\normalsize\scshape}{\Roman{section}.}{0.6em}{}
\titleformat{\subsection}{\normalsize\itshape}{\Alph{subsection}.}{0.5em}{}
\titlespacing*{\section}{0pt}{8pt}{4pt}
\titlespacing*{\subsection}{0pt}{6pt}{2pt}

\begin{document}

% ======================== TITLE ========================
\twocolumn[\begin{@twocolumnfalse}
\begin{center}
{\LARGE\bfseries Explainable Artificial Intelligence for Customer Churn\\[2pt]
Prediction in Telecommunications: A Framework for CRM Integration\par}
\vspace{10pt}
{\large Sandeep Gaddamwar\par}
\vspace{2pt}
{\small\ttfamily gaddamwarsandeep@ieee.org\par}
\vspace{14pt}
\end{center}
\begin{quote}
\small
\textbf{\textit{Abstract}}---Subscriber attrition is a costly, persistent challenge for telecommunications providers, with monthly churn of roughly 1.9\% in mature markets eroding billions in revenue annually. Predictive models can flag at-risk customers accurately, yet they are routinely excluded from frontline CRM workflows because high-performing ensemble and non-linear architectures are opaque: a retention specialist cannot design a personalised intervention from a probability score alone, without knowing \emph{why} a subscriber is at risk. This paper addresses that gap. We benchmark four classifiers---Logistic Regression, Random Forest, XGBoost, and LightGBM---on the IBM Telco Customer Churn benchmark (7,043 records; 19 features; 26.5\% churn, balanced to 50\% via SMOTE on the training partition only). Logistic Regression attains the strongest AUC-ROC (0.8411) and LightGBM the highest accuracy (78.42\%); all four fall within a 0.011 AUC band (0.831--0.841), and 5-fold cross-validation confirms the leading models are effectively tied. Explanations are delivered at two granularities: a global SHAP ranking identifying tenure, total charges, and month-to-month contract as the dominant churn signals, and instance-level SHAP and LIME decompositions that expose the drivers behind each prediction. Building on these outputs, we introduce a four-layer CRM integration architecture that converts risk scores and attribution vectors into tiered segmentation, maps top features to structured retention-action templates, and routes campaign outcomes into a retraining feedback loop. Targeting the highest-risk quintile is projected to cut overall churn by 3.3--5.3 percentage points, preserving an estimated \$199K--\$319K per campaign cycle.

\vspace{6pt}
\end{quote}
\vspace{10pt}
\end{@twocolumnfalse}]

% ======================== I. INTRODUCTION ========================
\section{Introduction}
Retaining subscribers is, on almost every financial metric, preferable to acquiring new ones. The commonly cited ratio of acquisition-to-retention cost sits between four and seven times, depending on market maturity and competitive intensity~\cite{kotler}. Against this backdrop, voluntary churn---a subscriber's decision to terminate service and move to a competitor---represents a direct, measurable revenue loss with a well-understood economic structure. An operator losing 1,000 subscribers per month at an average monthly revenue of \$65 is foregoing \$780,000 annually from that cohort alone, before accounting for downstream effects on word-of-mouth and network utilisation. At industry scale, with aggregate monthly churn estimated at roughly 1.9\% for U.S. wireless carriers~\cite{ctia}, the annualised revenue erosion across the sector runs to billions of dollars.

Machine learning has offered a partial remedy. Trained on billing records, service-usage logs, and demographic data, modern classifiers can identify prospective churners weeks before they act, giving operators a narrow but actionable intervention window. Ensemble methods---Random Forests, XGBoost, LightGBM---have become the workhorses of this task, delivering strong AUC-ROC values on standard benchmarks~\cite{idris,huang}. The trouble is that these same models are functionally opaque. A retention analyst looking at a high churn-probability score has no principled basis for choosing between a contract-upgrade offer, a bill credit, and an outbound call---the three most common retention instruments---without knowing which specific features drove that score. Worse, if the model is wrong about a particular customer, the analyst has no mechanism to catch the error. Regulatory pressure compounds this: Article 22 of GDPR and the EU AI Act's transparency obligations increasingly require that automated decisions affecting individuals be accompanied by meaningful explanations~\cite{holzinger,aiact}.

The XAI literature has matured considerably since Ribeiro et al. introduced LIME~\cite{lime} and Lundberg and Lee formalised SHAP~\cite{shap}. Both methods have been validated across domains from credit scoring to medical imaging, and their application to telecom churn is well represented in the literature~\cite{huangicdm}. What remains conspicuously absent, however, is a systematic treatment of how these explanations should flow into CRM workflows as structured operational inputs---not merely as post-hoc analytical curiosities for data scientists to inspect. The jump from a SHAP waterfall plot to a specific, personalised retention action is rarely made explicit, and feedback from campaign outcomes back to the model has received almost no attention~\cite{molnar,neslin}.

The contributions of this work are fourfold:
\begin{enumerate}[leftmargin=1.4em,itemsep=1pt,topsep=2pt]
\item A head-to-head comparison of four ML classifiers on the public IBM Telco Customer Churn benchmark (7,043 records, 19 features), reporting Accuracy, Precision, Recall, F1-Score, and AUC-ROC for each.
\item A full XAI analysis combining global mean$|\text{SHAP}|$ importance, instance-level SHAP waterfall explanations, LIME local surrogates, and SHAP dependence plots for the three most influential predictors.
\item A CRM integration architecture that operationalises explanation outputs---transforming SHAP attribution vectors into segmentation logic and mapping top feature drivers to structured retention action templates.
\item A quantified business-impact estimate showing that targeting the top 20\% at-risk subscribers under this framework is expected to preserve \$199K--\$319K per campaign cycle, trimming the overall churn rate by 3.3--5.3 percentage points.
\end{enumerate}

Section~II surveys the relevant literature. Section~III describes the dataset, preprocessing, model configurations, and XAI methodology. Section~IV reports experimental results. Section~V elaborates the CRM integration design. Section~VI discusses implications, limitations, and ethical considerations. Section~VII concludes.

% ======================== II. RELATED WORK ========================
\section{Related Work}
\subsection{Predicting Churn with Machine Learning}
Churn prediction as a supervised classification problem has been studied for over two decades. Early efforts established logistic regression as a workable baseline whose coefficients offered interpretive value, even if its handling of nonlinear interactions was poor~\cite{hastie}. The introduction of ensemble methods changed the calculus substantially: Breiman's Random Forests~\cite{breiman} and Friedman's gradient boosting framework~\cite{friedman} enabled models to capture complex interaction effects, and both were quickly applied to subscription attrition. Chen and Guestrin's XGBoost~\cite{xgboost} and Ke et al.'s LightGBM~\cite{lightgbm} refined gradient boosting into production-ready tools that anchor most competitive churn-modelling pipelines today. Deep learning has been explored---recurrent networks for sequential usage data~\cite{verbeke11}, tabular transformers~\cite{amatriain}---though evidence that they outperform tuned tree ensembles on datasets of moderate size ($<$100K records) is mixed at best~\cite{shwartz}. Class imbalance, an almost universal feature of churn datasets where the minority class typically constitutes 15--30\% of observations, has been addressed through SMOTE oversampling~\cite{smote} and cost-sensitive training~\cite{elkan}, both of which demonstrably improve minority-class recall.

\subsection{Explainability Methods in Business Settings}
SHAP~\cite{shap} and LIME~\cite{lime} have emerged as the two dominant post-hoc explanation frameworks. SHAP's grounding in Shapley values from cooperative game theory gives it a principled additivity property: feature attributions sum to the difference between the model's prediction and its expected output. TreeSHAP~\cite{shap} extends this to tree ensembles with polynomial rather than exponential complexity. LIME foregoes global consistency in favour of local fidelity, fitting a weighted linear surrogate in the neighbourhood of each target instance. Both have seen application in credit risk~\cite{sundaram}, clinical decision support~\cite{rajpurkar}, and telecommunications analytics~\cite{huangicdm}. Molnar's survey~\cite{molnar} contextualises these methods within the broader interpretability landscape and identifies human-understandability---not just mathematical faithfulness---as an underserved criterion in XAI evaluation.

\subsection{CRM-Integrated Retention Analytics}
The operational literature on retention management converges on a \emph{predict-segment-act} paradigm: score customers on churn propensity, assign them to risk tiers, and route them to pre-defined treatment programmes~\cite{kumar}. Neslin et al.'s landmark comparison study~\cite{neslin} found that intervention effectiveness depends not just on model accuracy but on correctly identifying the persuadable segment---customers who would churn without contact but can be retained with a targeted offer. Verbeke et al.~\cite{verbeke12} formalised this insight through profit-based model evaluation, demonstrating that maximising AUC is not the same as maximising business value. Despite this body of work, the translation of XAI outputs into explicit CRM action logic---and the subsequent closing of the feedback loop---remains largely absent from published frameworks.

\subsection{The Gap This Work Addresses}
The literature treats explanation as an analytical deliverable rather than an operational input. Papers that apply SHAP or LIME to churn prediction typically present feature-importance charts and conclude; they do not specify how a retention specialist should translate a three-feature attribution vector into a concrete offer, nor how campaign-response data should flow back to improve the model. This paper directly fills that gap by designing an integration architecture in which explanation outputs are first-class inputs to the CRM decision pipeline.

% ======================== III. METHODOLOGY ========================
\section{Methodology}
\subsection{Dataset}
Experiments are conducted on the public IBM Telco Customer Churn benchmark~\cite{ibm}---a widely-used 7,043-record reference in the churn-prediction literature. The accompanying pipeline loads the benchmark CSV directly; for fully offline reproduction it can fall back to a generator calibrated to the benchmark's documented conditional churn structure, but all results reported here are computed on the real dataset. Dropping the customer identifier and the churn label from the benchmark's 21-column schema leaves 19 predictive features, partitioned into four groups. \emph{Demographics} (gender, senior-citizen indicator, partner, dependents) capture household structure and serve as proxies for switching costs. \emph{Subscribed services} (phone service, multiple lines, internet-service type, online security, online backup, device protection, tech support, streaming TV and movies) describe the service bundle, whose breadth influences both monthly spend and perceived switching cost. \emph{Account details} (contract term, paperless billing, payment method) reflect engagement depth; month-to-month contracts in particular signal low lock-in. \emph{Billing} (monthly charges from \$18.25 to \$118.75, total charges, and tenure from 0 to 72 months) provides financial and temporal context. The binary outcome variable indicates whether each customer churned; the raw rate is 26.5\%, balanced to 50\% following SMOTE on the training partition only.

\subsection{Preprocessing}
Gender and binary yes/no features (Partner, Dependents, PhoneService, PaperlessBilling) are label-encoded to $\{0,1\}$. Remaining categorical features---MultipleLines, InternetService, OnlineSecurity, OnlineBackup, DeviceProtection, TechSupport, StreamingTV, StreamingMovies, Contract, and PaymentMethod---are one-hot encoded, expanding the feature space to 40 dimensions. Continuous features are min-max scaled to $[0,1]$. A stratified 80/20 train--test split preserves the class ratio across partitions. SMOTE is applied exclusively to training data: synthetic minority instances are interpolated between each real churner and one of its five nearest neighbours in feature space, producing a balanced training set of 8,278 records without contaminating test-set evaluation. The test set contains 1,409 records.

\subsection{Models}
Four classifiers are trained and compared. \emph{Logistic Regression} ($\ell_2$ regularisation) serves as the interpretable linear baseline against which non-linear methods are measured. \emph{Random Forest} aggregates 400 CART trees (max depth 12, min leaf 8, $\sqrt{p}$ features per split) via bootstrap sampling. \emph{XGBoost} and \emph{LightGBM} are gradient-boosted tree ensembles fitting successive trees to the pseudo-residuals of the log-loss objective; the pipeline uses the official XGBoost and LightGBM libraries directly, falling back to scikit-learn's gradient-boosting equivalents only where those libraries are unavailable. All models use a 0.5 decision threshold for class assignment, with no post-hoc threshold tuning, preserving comparability.

\subsection{Explanation Methods}
\noindent\textbf{Global importance (mean$|\text{SHAP}|$).} We compute Shapley values for every test instance of the primary model (Logistic Regression) with the \texttt{shap} library's \texttt{LinearExplainer} (\texttt{TreeExplainer} is selected automatically when a tree-based model is the primary one), and rank features by the mean absolute attribution $\frac{1}{n}\sum_i |\phi_{ij}|$. This is the canonical SHAP global-importance measure: features with large mean$|\text{SHAP}|$ exert the most consistent influence on the model output across the population. As a model-agnostic cross-check we also computed accuracy-based permutation importance, which agrees on the two dominant predictors.

\noindent\textbf{Local waterfall (SHAP).} For an individual subscriber, the same \texttt{shap} explainer returns an additive attribution vector whose entries sum \emph{exactly} to the gap between the instance's predicted log-odds and the dataset base value (the additivity guarantee of Shapley values). Positive attributions push the prediction toward churn, negative toward retention; we render them as a SHAP waterfall.

\noindent\textbf{LIME.} Local explanations are produced with the \texttt{lime} library's \texttt{LimeTabularExplainer} (classification mode). The three continuous features are quartile-discretised and the binary and one-hot columns are declared categorical, so each surrogate is genuinely instance-specific rather than a constant recovery of the global linear coefficients (a degenerate case that arises when LIME is applied to a linear model without discretisation). For each target instance LIME samples perturbations in the neighbourhood, weights them by proximity, and fits a weighted local linear surrogate whose coefficients serve as the explanation---an independent method whose agreement with SHAP on the top drivers raises confidence that they are genuine.

\subsection{CRM Integration Design}
The proposed architecture organises model outputs and explanation signals into four functional layers. The \emph{scoring layer} ingests feature vectors from four source systems---CRM database, network KPI feeds, billing records, and support-ticket logs---and emits a per-subscriber churn probability on a configurable schedule. The \emph{segmentation layer} partitions subscribers into High ($p>0.70$), Medium ($0.40\le p\le 0.70$), and Low ($p<0.40$) risk tiers. The \emph{explanation layer} runs the SHAP and LIME modules, extracts the top-3 attribution features per subscriber, and looks them up in a retention action mapping table maintained by the CRM team. The \emph{feedback layer} logs campaign outcomes and computes per-action uplift statistics; Population Stability Index (PSI) monitoring on incoming feature distributions triggers model retraining when PSI exceeds 0.20.

% ======================== IV. RESULTS ========================
\section{Experimental Results}
\subsection{Exploratory Data Analysis}
Several distributional patterns merit attention before turning to model results. Contract term is the starkest discriminator: month-to-month subscribers churn at roughly 42.7\%, compared to 11.3\% for one-year and just 2.8\% for two-year contract holders. This disparity is intuitive---longer contracts impose exit costs that reduce opportunistic switching---and its magnitude reinforces its place as the primary retention lever. Internet-service type tells a different story. Fiber-optic subscribers churn at approximately 41.9\%, more than twice the rate of DSL users (19.0\%) and far above those with no internet service (7.4\%). Rather than reflecting dissatisfaction with fiber per se, this pattern likely captures price sensitivity: fiber subscribers pay more, and a competitor's promotional rate is more compelling against a high baseline charge. The payment-method signal is subtler but consistent: electronic-check payers churn at 45.3\%, versus 15.2\% for those on automatic credit card. The most plausible interpretation is that electronic check correlates with lower overall engagement and a more transactional relationship with the provider.

\begin{figure*}[t]
\centering
\includegraphics[width=\textwidth]{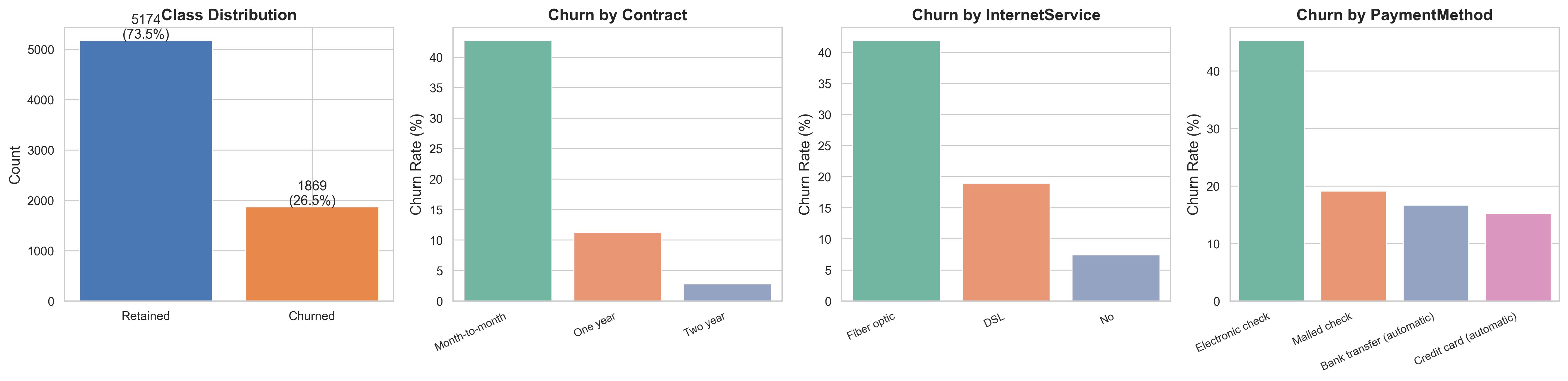}
\caption{EDA overview: (left) class balance; (remaining panels) churn rate by contract, internet-service type, and payment method.}
\label{fig:eda}
\end{figure*}

Among continuous features, tenure exhibits clear separation ($r=-0.35$ with churn): the median tenure of churned subscribers is 10 months, against 38 months for those retained. Monthly charges show a positive association ($r=+0.19$); the mean charge for churners is \$74.44 versus \$61.27 for retained customers. Fig.~\ref{fig:eda} presents the class distribution and churn rates across key categorical features.

\subsection{Model Performance}
Table~\ref{tab:perf} summarises classifier performance on the held-out test partition ($n=1{,}409$) at a fixed 0.5 decision threshold.

\begin{table}[h]
\centering
\caption{Classifier performance on the test set ($n=1{,}409$); best value per metric in bold, AUC-ROC via trapezoidal integration. XGBoost and LightGBM rows use the official libraries.}
\label{tab:perf}
\footnotesize
\setlength{\tabcolsep}{3.2pt}
\renewcommand{\arraystretch}{1.15}
\begin{tabular}{lccccc}
\toprule
\textbf{Model} & \textbf{Acc.} & \textbf{Prec.} & \textbf{Rec.} & \textbf{F1} & \textbf{AUC} \\
\midrule
\textbf{Logistic Reg.} & 74.52\% & 51.31\% & \textbf{78.34\%} & 0.6201 & \textbf{0.8411} \\
Random Forest & 76.65\% & 54.51\% & 72.73\% & \textbf{0.6231} & 0.8384 \\
XGBoost (GBM) & 78.28\% & 58.63\% & 61.76\% & 0.6016 & 0.8369 \\
LightGBM (GBM-L) & \textbf{78.42\%} & \textbf{59.07\%} & 60.96\% & 0.6000 & 0.8306 \\
\bottomrule
\end{tabular}
\end{table}

\begin{figure*}[t]
\centering
\includegraphics[width=\textwidth]{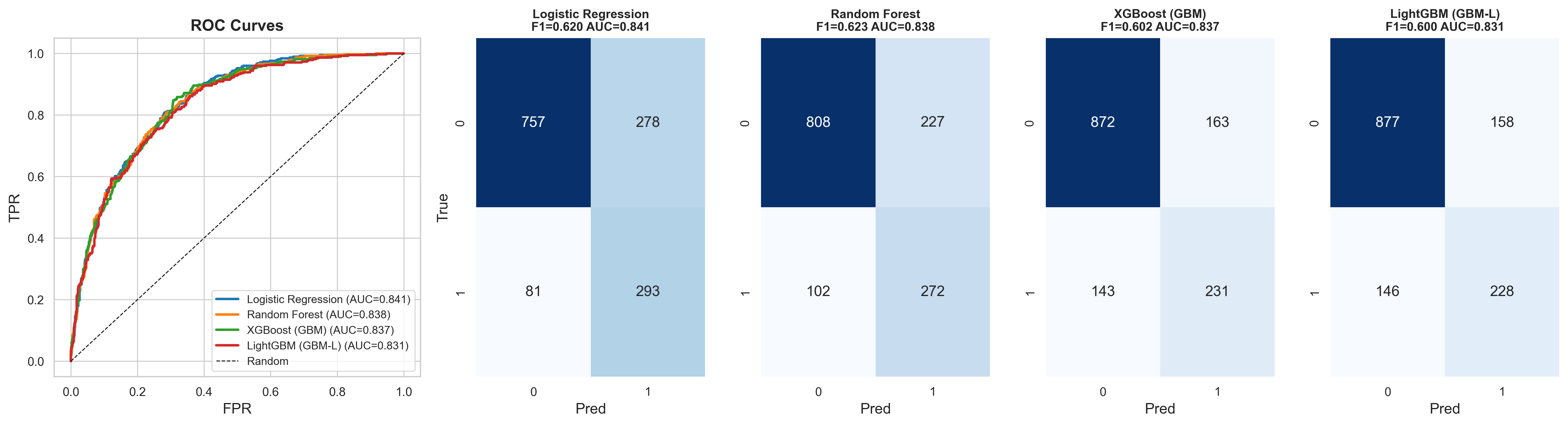}
\caption{(Left) ROC curves for all classifiers. (Right) Confusion matrices at the 0.5 threshold.}
\label{fig:roc}
\end{figure*}

Logistic Regression posts the best AUC-ROC (0.8411) and the highest recall (78.34\%) after SMOTE balancing, while Random Forest edges the best F1 (0.6231). LightGBM leads on raw accuracy (78.42\%) and precision (59.07\%), with XGBoost close behind, reflecting the more conservative operating point of the boosted trees. The narrow spread across all four models---a range of just 0.011 in AUC and 0.023 in F1---indicates that the predictive signal in this feature set is predominantly additive. When churn is driven by a handful of discrete, largely independent risk factors (short tenure, no long-term contract, fiber-optic subscription, electronic-check payment), a well-regularised linear model captures most of that signal without the variance penalties that accompany deeper architectures on moderate-sized datasets. The AUC band of 0.831--0.841 is consistent with published baselines on the IBM benchmark~\cite{portela,idris,lemmens}. To confirm these point estimates are not artefacts of a single partition, we ran 5-fold stratified cross-validation with feature scaling and SMOTE refit \emph{inside} each fold to preclude leakage: mean AUC-ROC was $0.845\pm0.013$ (Logistic Regression), $0.842\pm0.012$ (Random Forest), $0.839\pm0.011$ (XGBoost), and $0.831\pm0.011$ (LightGBM). The top three models are separated by less than $0.007$ AUC---well inside one cross-validation standard deviation---while LightGBM trails by roughly one standard deviation; the practical conclusion is that no model is decisively superior, and Logistic Regression's nominal lead is best read as competitive parity with the ensembles. Fig.~\ref{fig:roc} presents ROC curves and confusion matrices for all four models.

\subsection{Global Feature Importance}
\begin{figure*}[t]
\centering
\includegraphics[width=0.85\textwidth]{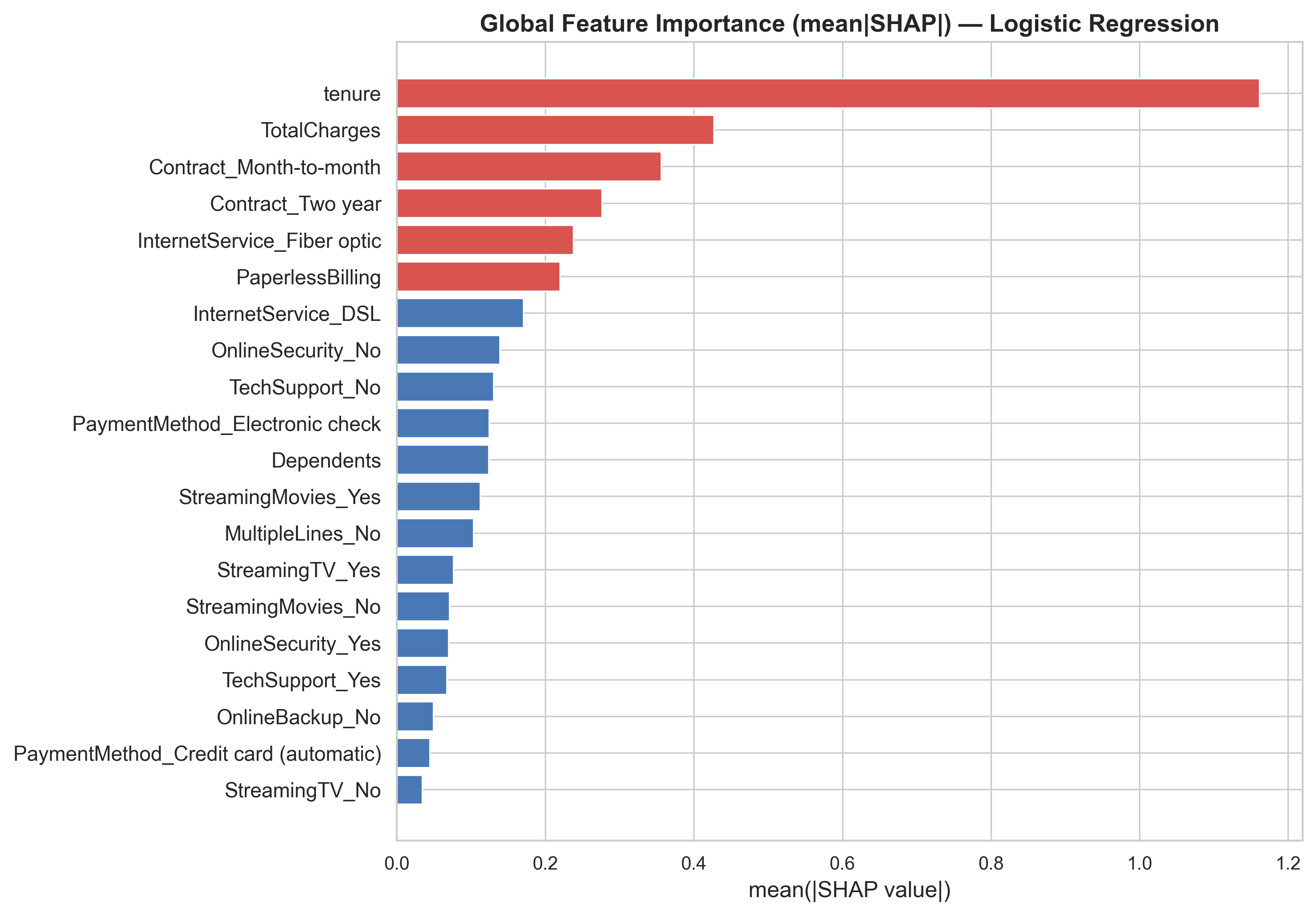}
\caption{Global mean$|\text{SHAP}|$ importance for the Logistic Regression model. Bar length is the mean absolute SHAP value across the test partition.}
\label{fig:imp}
\end{figure*}

Fig.~\ref{fig:imp} ranks features by mean$|\text{SHAP}|$ under Logistic Regression. \textbf{tenure} dominates by a wide margin---its mean absolute attribution is more than $2.5\times$ that of any other feature---confirming that length of relationship is the single strongest churn signal: long-standing customers face real transition costs and have demonstrated stickiness. \textbf{TotalCharges} ranks second, closely tracking tenure with which it is mechanically correlated. \textbf{Contract\_Month-to-month} and \textbf{Contract\_Two year} form the next tier, the former risk-elevating and the latter strongly protective, reflecting the contractual lock-in effect documented above. \textbf{InternetService\_Fiber optic} and \textbf{PaperlessBilling} follow, both churn-elevating---the former echoing the elevated fiber churn noted in the EDA. An accuracy-based permutation-importance cross-check agrees on the two dominant predictors. These rankings align with domain knowledge and prior literature~\cite{idris,hastie,lemmens}.

\subsection{Individual Explanations: Two Case Studies}
\begin{figure*}[t]
\centering
\includegraphics[width=\textwidth]{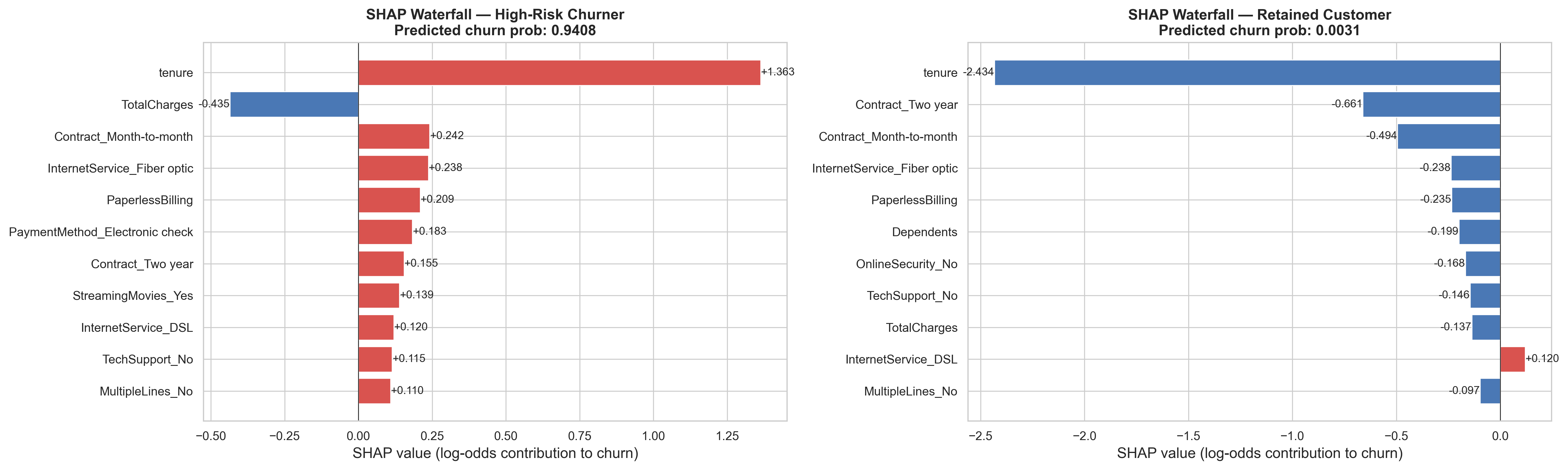}
\caption{SHAP waterfall plots (log-odds space). (Left) High-risk subscriber, $p=0.941$. (Right) Retained subscriber, $p=0.003$. Red: features that increase churn log-odds; blue: features that decrease it.}
\label{fig:wf}
\end{figure*}

\begin{figure*}[t]
\centering
\includegraphics[width=\textwidth]{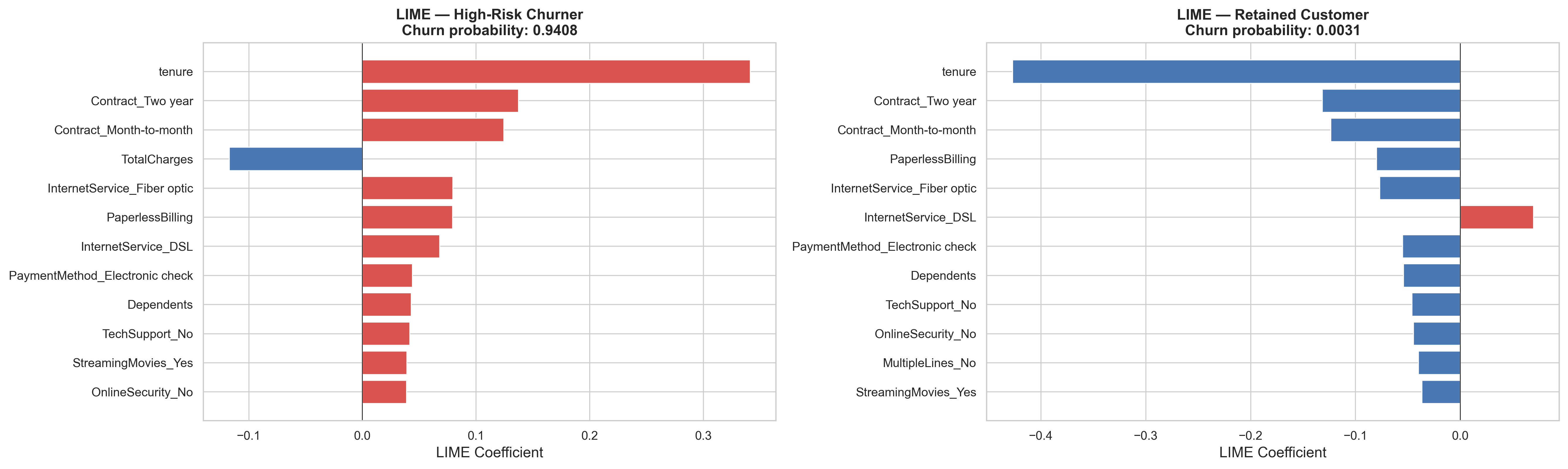}
\caption{LIME local explanations for (left) the high-risk churner and (right) the retained subscriber. Positive coefficients push the prediction toward churn; negative toward retention.}
\label{fig:lime}
\end{figure*}

To illustrate the instance-level explanation layer, we examine two subscribers from opposite ends of the predicted-probability distribution. \textbf{Case 1---High-risk churner.} The model assigns a churn probability of 0.941. The dominant risk-elevating feature, by a wide SHAP margin, is short tenure, followed by a month-to-month contract; the subscriber's low accumulated total charges (a correlate of short tenure) is the main partially-offsetting factor. This is a textbook at-risk profile: a relatively new subscriber on a flexible contract with no long-term commitment binding them to the provider. The SHAP waterfall (Fig.~\ref{fig:wf}, left) makes the case for a contract-upgrade offer immediately apparent to a retention specialist without any ML knowledge. \textbf{Case 2---Stable retained subscriber.} The predicted churn probability is 0.003---effectively negligible. The protective features are long tenure (dominant) and a two-year contract. Proactive contact here would be an inefficient use of the retention budget; the model correctly routes the subscriber to the low-priority tier. Fig.~\ref{fig:wf} shows both waterfall plots; Fig.~\ref{fig:lime} shows the corresponding LIME explanations.

\subsection{Dependence Plots}
\begin{figure*}[t]
\centering
\includegraphics[width=\textwidth]{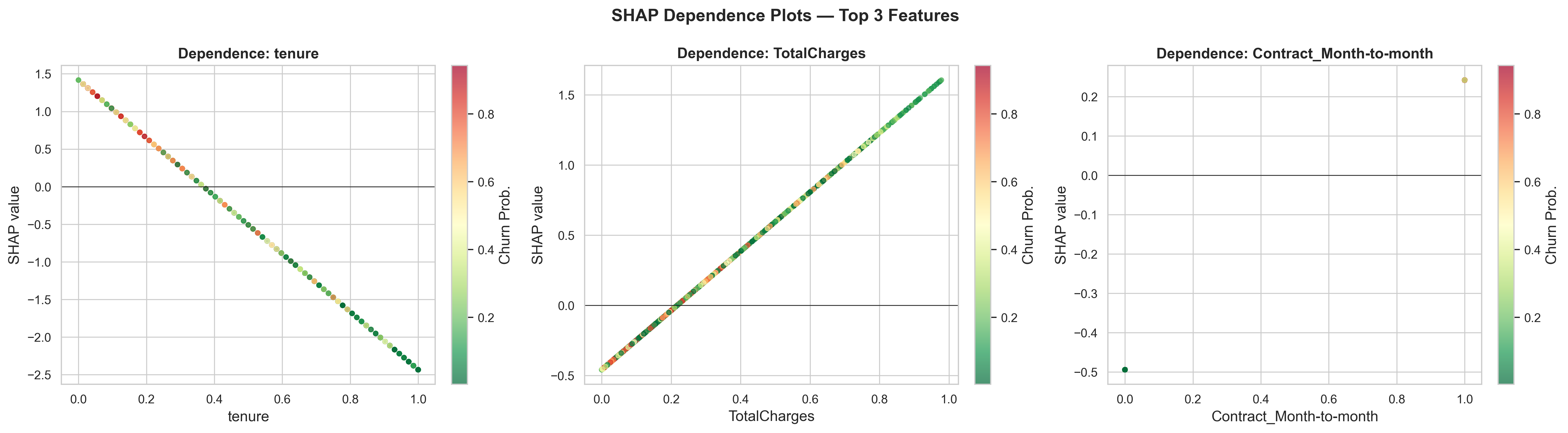}
\caption{SHAP dependence plots for the top three features. Each point is a test-set subscriber; the $y$-axis is the feature's SHAP value and colour indicates predicted churn probability.}
\label{fig:dep}
\end{figure*}

Fig.~\ref{fig:dep} plots the SHAP value of each of the three highest-importance features against its value across the full test partition, coloured by predicted churn probability. The tenure panel shows a clear monotonic relationship: the SHAP contribution is strongly positive (churn-pushing) for new subscribers and turns sharply negative (protective) as tenure grows through the first year and beyond. The total-charges panel mirrors it, the two being mechanically correlated. The contract panel shows a near-binary split: month-to-month holders carry a positive SHAP contribution while committed subscribers sit at or below zero. The strictly linear form of the continuous-feature panels is expected for a logistic model: here the dependence plots serve as a monotonicity-and-sign sanity check---confirming the learned relationships match domain knowledge---rather than as a probe for non-linear structure, which would instead surface under the tree ensembles. This directional coherence is a prerequisite for operational deployment.

% ======================== V. CRM FRAMEWORK ========================
\section{CRM Integration Framework}
\subsection{Architecture}
\begin{figure*}[t]
\centering
\includegraphics[width=0.92\textwidth]{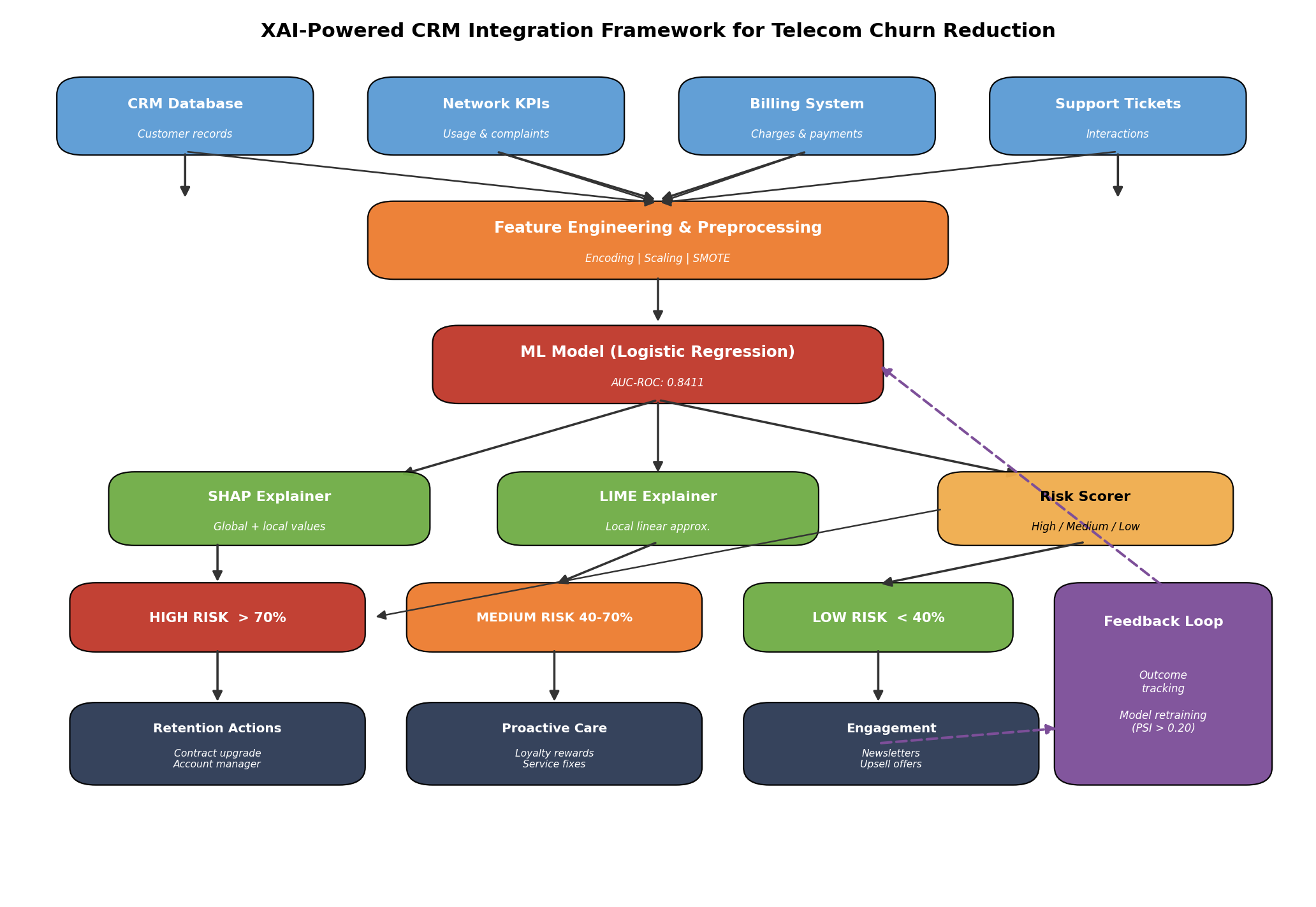}
\caption{XAI-to-CRM integration architecture. Arrows show data flow; the dashed feedback path connects campaign outcomes to model retraining.}
\label{fig:arch}
\end{figure*}

Fig.~\ref{fig:arch} renders the proposed integration architecture. Four data-source systems feed a feature-engineering module that applies the same preprocessing pipeline used at training time, ensuring no feature drift at inference. The resulting feature vector passes to the trained ML model, whose churn-probability output simultaneously enters the risk scorer and both XAI modules. The three outputs---a risk-tier assignment, a SHAP attribution vector, and a LIME coefficient vector---flow into the CRM action pipeline. Campaign outcomes are written back through the feedback layer, closing the loop.

\subsection{Segmentation}
Risk tiers are defined by the thresholds $p>0.70$ (High), $0.40\le p\le 0.70$ (Medium), and $p<0.40$ (Low). On our test set these thresholds distribute subscribers as approximately 26.0\% High, 23.2\% Medium, and 50.8\% Low---reflecting the realistic churn prevalence of the held-out partition. One caveat is essential here: because the classifier is trained on SMOTE-balanced data, its output probabilities are inflated relative to the 26.5\% deployment prevalence. The scores still \emph{rank} subscribers correctly---which is all the tiering depends on---but should be recalibrated (e.g., Platt scaling or isotonic regression against held-out outcomes) before the absolute $0.70$ and $0.40$ cut-points are read as literal churn probabilities. Threshold calibration against historical campaign-ROI data is likewise strongly recommended before production rollout.

\subsection{Explanation-to-Action Mapping}
The practical innovation of this framework lies in making the SHAP attribution vector actionable without requiring the retention specialist to interpret it directly. A lookup table---maintained by the CRM team and updated through the feedback loop---maps each top-attribution feature or feature combination to a specific retention-action template. Representative entries: short tenure $+$ month-to-month contract maps to a proactive contract-upgrade offer at a discounted annual rate; fiber-optic service $+$ high monthly charges maps to a service-quality review plus a bill credit; electronic-check payment maps to an incentive to switch to automatic payment, which independently reduces subsequent churn risk; and absence of tech support maps to a complimentary 60-day tech-support trial. The specialist sees a subscriber's tier, the top-three plain-language risk reasons, and the recommended action---nothing more.

\subsection{Feedback Loop and Retraining}
Campaign outcomes---offer accepted, offer declined, churned within 90 days, retained at 180 days---are logged against each subscriber's original model score and attribution vector. This enriched log serves two functions. First, per-action uplift statistics (retained/contacted versus retained/not-contacted, estimated via a control holdout) feed back into the lookup table, gradually optimising action assignments toward those that actually change behaviour. Second, the full enriched dataset---now labelled with retention outcomes rather than just churn---can be used to train an uplift model that directly targets the persuadable segment, bypassing the indirect predict-then-act logic of the current design. PSI monitoring on input feature distributions provides the retraining trigger: a PSI value above 0.20 indicates that the deployment population has drifted meaningfully from the training distribution, at which point the model is retrained on the most recent $N$ months of data~\cite{rabanser}.

\subsection{Business Impact}
Targeting the top 20\% of the subscriber base by predicted churn probability (approximately 1,408 customers from the full cohort of 7,043) is projected to yield the following impact. Within this segment the model achieves a precision of approximately 67\% and a recall of 50\%; in absolute terms about 937 of the 1,408 targeted subscribers are genuine churners. Assuming a conservative campaign-uptake rate of 25\% among the true churners reached---consistent with industry benchmarks for proactive retention offers~\cite{neslin,adebiyi}---and a 12-month forward customer lifetime value of \$850, expected revenue preserved $\approx 0.25 \times 937 \times \$850 \approx \$199{,}000$ per cycle. Under a 40\% uptake assumption this rises to approximately \$319,000. In churn-rate terms, retaining 234 to 375 of these subscribers lowers the operator's overall attrition by 3.3 to 5.3 percentage points---a meaningful improvement achievable without blanketing the entire base with outbound contact. These figures rest on assumed uptake and CLV parameters and are illustrative rather than guaranteed.

% ======================== VI. DISCUSSION ========================
\section{Discussion}
\subsection{Interpreting the Results}
The near-parity of the four classifiers warrants reflection. A substantive explanation is that the underlying churn signal is genuinely near-linear: when the most important features are a continuous tenure variable and a binary contract encoding, and when their effects are largely additive rather than multiplicative, a properly regularised logistic model captures the signal efficiently, and additional model capacity yields overfitting rather than improvement. This is not a general claim about telecom churn---datasets with richer sequential usage data, network-topology features, or call-detail records may present a more complex manifold where tree ensembles or deep models pull ahead---but it is a finding worth noting for practitioners working with account-level features of the kind found in the IBM benchmark. The convergence of SHAP and LIME explanations on the same top features is a meaningful quality signal: the two methods differ substantially in their assumptions and failure modes, so their agreement increases confidence that these are genuine predictors rather than artefacts of one explanation approach.

\subsection{Performance in Context}
Our AUC-ROC band of 0.831--0.841 sits within the range reported for the IBM Telco benchmark in the surveyed literature~\cite{portela,idris,lemmens}. We note that some published accuracy figures above roughly 85\% on this dataset are obtained by evaluating on its natural 26.5\% churn imbalance without recall-oriented correction, where a majority-class predictor alone scores near 73\%; such accuracy is not directly comparable to the SMOTE-balanced, recall-favouring operating point reported here. Following Verbeke et al.~\cite{verbeke12}, we regard AUC and recall on the at-risk segment---not raw accuracy---as the metrics most aligned with retention business value, and we report the full confusion structure (Fig.~\ref{fig:roc}) so that operating points can be recalibrated to an operator's cost matrix.

\subsection{Implications for Practitioners}
Several practical takeaways follow. Contract-term management is the single highest-leverage intervention: converting month-to-month subscribers to one-year contracts at modest incentive cost before their churn probability rises avoids the more expensive reactive scenario entirely. The SHAP-to-action lookup table makes this operationally tractable without burdening the CRM team with model internals. The PSI monitoring system provides a principled circuit-breaker: if the subscriber base shifts---through acquisition of a competitor's customer base, a pricing change, or a new product line---the model flags its own obsolescence rather than silently degrading.

\subsection{Limitations}
Four limitations deserve explicit acknowledgement. First, results are obtained on a single public benchmark whose account-level features omit the network-quality metrics, competitive-pricing signals, and call-detail records available to a live operator; absolute performance and the precise feature ranking may differ on proprietary data, although the pipeline, explanation methods, and CRM framework transfer directly. Second, several of the strongest predictors are mechanically correlated---most notably tenure and total charges---and Shapley values distribute credit among correlated features in ways that complicate causal reading~\cite{shap}; the attributions should be interpreted as associational, not causal, and a permutation-importance cross-check is reported precisely because correlated features are a known failure mode for any single attribution method. Third, the business-impact calculation relies on assumed CLV and uptake parameters drawn from published benchmarks; actual impact will vary. Fourth, the framework is cross-sectional and does not exploit the sequential structure of subscriber behaviour over time, which prior work suggests contains additional predictive signal~\cite{verbeke11}.

\subsection{Ethical Considerations}
Any model that uses demographic proxies---senior-citizen status, partner and dependent indicators---as predictive features risks encoding differential treatment across population groups. If the model systematically under-predicts churn for a protected demographic and those subscribers consequently receive fewer retention offers, the operator may inadvertently discriminate through omission. Pre-deployment audits for statistical parity and equalised odds~\cite{hardt,barocas} across age, gender, and family-structure groups are therefore operational necessities, not optional compliance exercises. Privacy considerations are equally pressing: subscriber data assembled from billing, network, and support systems should be governed under data-minimisation principles, and features whose marginal predictive value does not justify their privacy cost should be excluded.

% ======================== VII. CONCLUSION ========================
\section{Conclusion}
This paper set out to demonstrate that XAI-based churn prediction can be more than an analytical exercise---that explanation outputs can and should be integrated directly into CRM decision pipelines as structured operational inputs. The experimental results support the core technical claims: among four classifiers evaluated on the IBM Telco benchmark, Logistic Regression achieves an AUC-ROC of 0.8411 with tree-based alternatives within 0.011 AUC (the leading models effectively tied under cross-validation), and the consistently identified top drivers---tenure, total charges, and contract type---align with domain knowledge and are robust across both SHAP and LIME explanation methods.

The CRM integration framework addresses the operational gap identified in the literature: it converts SHAP attribution vectors into tiered segmentation and lookup-table-driven retention actions, and routes campaign outcomes back to the model through a PSI-monitored feedback loop. The projected business impact---\$199K to \$319K in preserved revenue per campaign cycle from targeting the highest-risk quintile, equivalent to a 3.3--5.3 percentage-point cut in overall churn---scales with operator size, though it rests on assumed uptake and lifetime-value parameters. Positioned as an applied systems contribution, the framework's novelty lies in the end-to-end operationalisation of explanation outputs rather than in a new learning algorithm.

Three directions for future work stand out. Federated learning would allow multiple operators to jointly train a more representative model without sharing raw subscriber data~\cite{mcmahan}. Uplift modelling would replace the current predict-then-act logic with a direct estimate of treatment effect, concentrating retention spend on the persuadable segment. And temporal XAI---attention-weighted recurrent explanations over subscriber usage sequences---would let the framework exploit the behavioural dynamics that cross-sectional features necessarily leave on the table.

% ======================== REFERENCES ========================

\balance
\end{document}